\documentclass[runningheads]{llncs}
\usepackage[T1]{fontenc}
\usepackage{amsmath}
\usepackage{hyperref}
\usepackage{graphicx,verbatim}
\usepackage{booktabs}  
\usepackage{multirow}  
\usepackage{makecell}
\usepackage{amssymb}   
\usepackage{siunitx}
\begin{document}
\title{Synergizing Discriminative Exemplars and Self-Refined Experience for MLLM-based In-Context Learning in Medical Diagnosis}
%

\author{
Wenkai Zhao\inst{1} \and
Zipei Wang\inst{1} \and
Mengjie Fang\inst{1} \and
Di Dong\inst{1, *} \and
Jie Tian\inst{1, *} \and
Lingwei Zhang\inst{2}
}

\authorrunning{Anonymized Author et al.}

\institute{
CAS Key Laboratory of Molecular Imaging, Institute of Automation, Chinese Academy of Sciences \and
School of Software, North University of China
}
  
\maketitle              
\begin{abstract}
General Multimodal Large Language Models (MLLMs) often underperform in capturing domain-specific nuances in medical diagnosis, trailing behind fully supervised baselines. Although fine-tuning provides a remedy, the high costs of expert annotation and massive computational overhead limit its scalability. To bridge this gap without updating the weights of the pre-trained backbone of the MLLM, we propose a Clinician Mimetic Workflow. This is a novel In-Context Learning (ICL) framework designed to synergize Discriminative Exemplar Coreset Selection (DECS) and Self-Refined Experience Summarization (SRES). Specifically, DECS simulates a clinician's ability to reference "anchor cases" by selecting discriminative visual coresets from noisy data at the computational level; meanwhile, SRES mimics the cognition and reflection in clinical diagnosis by distilling diverse rollouts into a dynamic textual Experience Bank. Extensive evaluation across all 12 datasets of the MedMNIST 2D benchmark demonstrates that our method outperforms zero-shot general and medical MLLMs. Simultaneously, it achieves performance levels comparable to fully supervised vision models and domain-specific fine-tuned MLLMs, setting a new benchmark for parameter-efficient medical in-context learning. Our code is available at an anonymous repository:
\href{https://anonymous.4open.science/r/Synergizing-Discriminative-Exemplars-and-Self-Refined-Experience-ED74}{anonymous repository}.


\end{abstract}
\section{Introduction}
Multimodal Large Language Models (MLLMs) facilitate generalist medical AI through interpretable reasoning and conversational interaction, surpassing traditional "black-box" vision models\cite{1}. However, general-purpose MLLMs often underperform task-specific supervised baselines in medical classification\cite{2,3}. While developing domain-specific models via fine-tuning is a common remedy, the required annotated data and computational resources are often prohibitive\cite{4}. To circumvent these costs, In-Context Learning (ICL) has emerged as a promising alternative\cite{5}, enabling a frozen backbone MLLM to adapt to specialized medical tasks through task-specific demonstrations without parameter updates\cite{6}.

To harness this promise, recent studies have actively explored adapting frozen backbone MLLMs via ICL for medical image analysis. Methodologies such as Active Prompt Tuning (APT) \cite{7} and Active In-Context Learning (AICL) \cite{8} employ iterative hard sample mining and spectral clustering to curate Discriminative Visual Exemplars. In parallel, MMRAG \cite{9} utilizes external knowledge bases to guide analogical reasoning, while Iris \cite{10} and K-Prism \cite{11} extend ICL to the more challenging domain of segmentation via context task encoding and dual-prompt representations. Despite these strides in prompting strategies, frozen backbone MLLMs frequently still struggle with fine-grained diagnosis due to inherent architectural constraints. To mitigate this, recent works \cite{12} suggest overlaying visual markers to explicitly guide attention. However, such interventions necessitate labor-intensive manual annotation, limiting their scalability.

To address these challenges, instead of relying on labor-intensive external visual prompts, a more scalable solution is to stimulate the internal cognitive capabilities of MLLMs \cite{cai2025training,cheng2025vision,yang2025probabilistic}. Therefore, we propose a Clinician Mimetic Workflow that computationally emulates human diagnostic reasoning. Clinical expertise inherently relies on two indispensable dimensions: comparative analysis via Illness Scripts \cite{13} and epistemic reflection for hypothesis evaluation \cite{14}. Our core contribution lies in synergistically integrating these two cognitive dimensions into a unified framework. Specifically, Discriminative Exemplar Coreset Selection (DECS) builds a visual coreset to simulate distinctive case referencing, while Self-Refined Experience Summarization (SRES) evolves a textual Experience Bank to mimic reflective rule updating. This synergy equips general-purpose MLLMs with specialized precision. Our main contributions are summarized as follows:
\begin{enumerate}
    \item A Novel Clinician Mimetic Workflow: We propose a unified framework integrating both visual comparison and epistemic reflection, enabling MLLMs to achieve effective domain adaptation without any backbone parameter updates.
    \item Simulation of Clinical Diagnostic Learning: We introduce two synergistic methods, DECS and SRES, to mimic the diagnostic learning process of clinicians.
    \item Superior Diagnostic Performance: Systematic experiments across the 12 2D datasets of MedMNIST demonstrate that our method outperforms zero-shot general and medical MLLMs on the majority of tasks, while achieving competitive results on all datasets when compared to fully supervised vision baselines and fine-tuned medical MLLMs.
\end{enumerate}

\begin{figure}
\centering
\includegraphics[width=\textwidth]{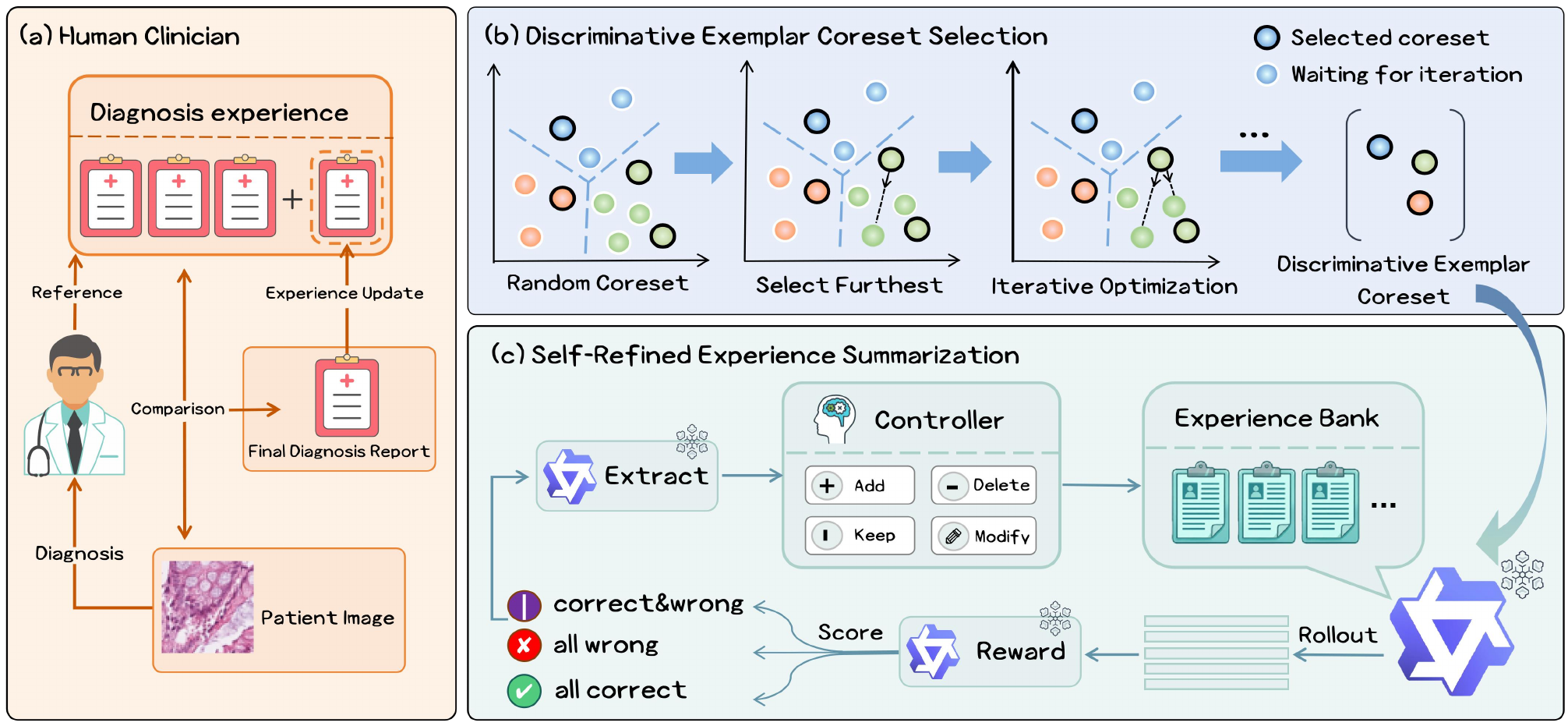} 
\caption{Overview of the proposed Clinician Mimetic Workflow. (a) The diagnostic cognitive process of human clinicians. (b) Discriminative Exemplar Coreset Selection, where data points of the same color belong to the same class. (c) Self-Refined Experience Summarization, dynamically updating the Experience Bank via reward-guided rollouts.} 
\label{fig:framework}
\end{figure}

\section{Methods}
\subsection{Overall Architecture}
Our Clinician Mimetic Workflow (Fig. 1) intrinsically augments a frozen backbone MLLM with visual structural priors and textual logical priors. It operates by seamlessly synergizing two core components: DECS and SRES. DECS computationally curates a visual coreset of distinctive cases from a labeled support set, providing dynamic visual comparative anchors for a given query. Concurrently, SRES evaluates generated reasoning paths via a reward-guided rollout mechanism, distilling diagnostic heuristics into an evolving Experience Bank. By synthesizing the DECS-retrieved visual exemplars with SRES-derived textual heuristics as holistic multimodal in-context prompts, our framework enables the frozen MLLM to perform grounded comparative analysis and deliver precise medical diagnoses without requiring any backbone parameter updates.

\subsection{Discriminative Exemplar Coreset Selection}

Due to the inherent intra-class variance in medical datasets, such as contrast shifts and varying staining intensities, traditional few-shot exemplar selection relying on nearest-neighbor matching is often sub-optimal. Our DECS attempts to average out irrelevant domain noise to extract discriminative exemplars.

\subsubsection{Feature Extracting and Adaptive Initialization}

Given a training set of size $N$, we extract $L_2$-normalized features $k_i=\frac{\phi(x_i)}{\lVert \phi(x_i)\rVert_2}$ using a visual encoder $\phi$. A class-balanced coreset $\mathcal{C}^{(0)}$ is initialized randomly. To balance capacity and efficiency, its size $S_{c}$ and optimization epochs $T_{opt}$ are adaptively scaled via $S_{c}=\lfloor B_{size}\sqrt{N/N_{ref}}\rfloor$ and $T_{opt}=\lfloor B_{epoch}\sqrt{N_{ref}/N}\rfloor$, where $B_{size}$, $B_{epoch}$, and $N_{ref}$ are base hyperparameters.

\subsubsection{Iterative Optimization of Distinctive Cases}

We iteratively optimize the keys $k \in \mathcal{C}$ within the feature space. Let the keys in the coreset for a specific class be indexed as $k^{(j)}$. For a batch of untapped query images $\mathcal{Q}$, we assign each query $q$ to its hardest positive prototype—that is, the least similar discriminative prototype index $j^*$ within the same class $y(q)$: 

\begin{equation}j^* = {argmin}_{\{j \mid k^{(j)} \in \mathcal{C}_{y(q)}\}} \langle \phi(q), k^{(j)} \rangle\end{equation}

This strategy forces the prototype to capture invariant disease features rather than superficial visual similarities. Subsequently, queries assigned to the same target index $j^*$ are aggregated into a subset $\mathcal{Q}^{(j^*)}$. We then utilize an exponential moving average (EMA) to update $k^{(j^*)}$ towards the semantic center of these dissimilar queries:

\begin{equation}k^{(j^*)} \leftarrow \text{norm}\left( (1 - \alpha) k^{(j^*)} + \alpha \frac{1}{\lvert \mathcal{Q}^{(j^*)} \rvert} \sum_{q \in \mathcal{Q}^{(j^*)}} \phi(q) \right)\end{equation}

where $\alpha$ is the update rate. The purpose of this intra-class hard-target aggregation is to average out uninformative visual variations (e.g., contrast changes or staining noise) and reinforce shared diagnostic biomarkers, thereby effectively mitigating the noise reinforcement and prototype drift issues prevalent in traditional similarity-based clustering.

\subsubsection{Inference-time Exemplar Referencing}

During inference, given a test image $x_{test}$, we compute its feature $\phi(x_{test})$ and reference the Top-$K$ Discriminative Visual Exemplars from $\mathcal{C}$ based on the cosine similarity with the optimized keys $k$. These exemplars are integrated with Diagnostic Heuristics from the Experience Bank.

\begin{figure}[!t]

\centering

\includegraphics[width=\textwidth]{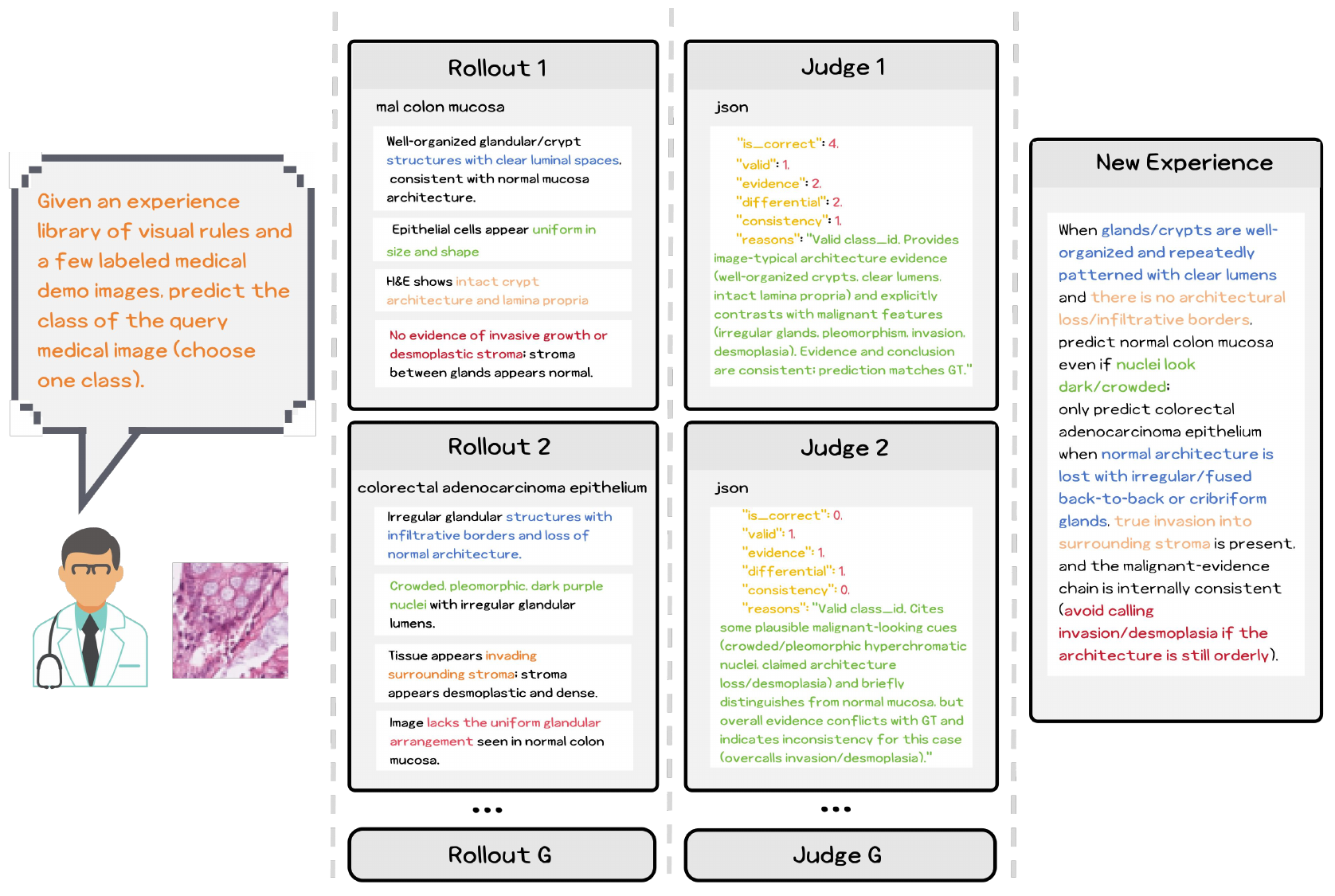} 

\caption{Illustration of the Self-Refined Experience Summarization. Texts highlighted in identical colors map the critical semantic discrepancies between distinct rollouts directly to the resulting synthesized Diagnostic Heuristics.} 

\label{fig:sres}

\end{figure}

\subsection{Self-Refined Experience Summarization}

As illustrated in Fig. \ref{fig:sres}, SRES dynamically distills clinical knowledge by evaluating diverse reasoning rollouts and synthesizing their semantic discrepancies.

\subsubsection{Contextual Formulation}

We define the Experience Bank as a dynamic set of textual heuristics $\mathcal{B}_{text} = \{h_m\}_{m=1}^M$. For a query image $x_q$, we integrate the DECS-referenced visual exemplars $\mathcal{C}_{topK}$. To guide inference, we construct a multimodal prompt $\mathcal{P}$ by concatenating ($\oplus$) the visual components via a formatting function $f_{vis}$ and the textual bank:

\begin{equation}\mathcal{P} = f_{vis}(x_q, \mathbf{\mathcal{C}_{topK}}) \oplus \text{Prompt}(\mathbf{\mathcal{B}_{text}})\end{equation}

This enables the frozen MLLM to concurrently leverage comparative visual anchors and accumulated expertise.

\subsubsection{Reward-Guided Rollout and Contrastive Summarization}

To mimic differential diagnosis, we sample a group of $G$ reasoning paths $\mathcal{R}=\{r_{g}\}_{g=1}^{G}$ from the MLLM policy $\pi(\cdot|\mathcal{P})$. An LLM evaluator $\mathcal{M}_{eval}$ then assigns a scalar reward $s_{g}$ to each path $r_{g}$ based on a predefined clinical criteria set $\Omega$ (correctness, validity, comprehensiveness, discriminability, consistency):

\begin{equation}s_{g}=\sum_{\omega_{j}\in\Omega}\mathcal{M}_{eval}(\omega_{j}|r_{g},\mathcal{P},y_{gt})\end{equation}

where $\omega_{j}$ represents an individual criterion and $y_{gt}$ is the ground truth. Crucially, $y_{gt}$ is strictly masked during the subsequent phase to prevent label leakage. The MLLM then scrutinizes the semantic gap between high-scoring $(r^{+})$ and low-scoring $(r^{-})$ trajectories within $\mathcal{R}$ to synthesize a concise diagnostic heuristic via $h_{new}=\text{Summarization}(\mathcal{R}, \{s_{g}\}_{g=1}^{G})$.

\subsubsection{Experience Bank Update}

To prevent saturation with redundant or low-quality information, we impose a maximum capacity $M_{max}$. When a new candidate rule $h_{new}$ is generated, the MLLM acts as an Evaluator to output an action $a^{(\tau)}$ from the update policy $\mathcal{T}$ at step $\tau$:

\begin{equation}a^{(\tau)} = \mathcal{T}(\mathcal{B}_{text}^{(\tau)}, h_{new}) \in \{ \text{Add}, \text{Delete}, \text{Keep}, \text{Modify} \}\end{equation}

The Experience Bank is then transitioned to the next state:

\begin{equation}\mathcal{B}_{text}^{(\tau+1)} = \text{Update}(\mathcal{B}_{text}^{(\tau)}, h_{new}, a^{(\tau)})\end{equation}

Specifically, the policy Adds novel insights, Deletes redundant, vague, or low-confidence entries, Keeps robust rules that remain highly relevant, or Modifies existing rules for precision and coverage.

\section{Experiments and Results}
\subsection{Experimental Settings}
\subsubsection{Datasets}
We conduct comprehensive evaluations on the 12 2D biomedical datasets from the public MedMNIST benchmark \cite{MedMNIST} (licensed under CC BY 4.0, except DermaMNIST under CC BY-NC 4.0). To preserve fine-grained medical features essential for clinical reasoning, we adopt the $224 \times 224$ resolution versions provided by the datasets. These datasets span diverse modalities including X-ray, Optical Coherence Tomography (OCT), Ultrasound, Computed Tomography (CT), Dermoscopy, Pathology, and Fundus Camera, covering binary, multi-class, and multi-label classification tasks.

\subsubsection{Implementation Details}
Experiments are implemented using PyTorch and vLLM on a single RTX 6000 GPU, adopting Qwen3-VL-8B as the frozen backbone. DECS utilizes a linear-probed SigLIP2 for feature extracting. The MLLM backbone remains strictly frozen. We optimize DECS on the full training set (batch size 128, update rate $\alpha = 0.2$, scaling parameters $N_{ref} = 1000$, $B_{size} = 50$, $B_{epoch} = 10$). The Experience Bank evolves efficiently using 200 random training instances (rollout $G=5$, temperature $T=1.0$, max capacity of 16 rules). A unified $N_{shot}=5$ referencing setting is adopted for all phases.

\subsubsection{Evaluation Metrics}
Given our focus on the correctness of final diagnostic decisions, we adopt Accuracy (ACC) as our primary evaluation metric. We acknowledge that the Area Under the ROC Curve (AUC) is a valuable metric. However, unlike discriminative models, MLLMs are generative models whose probability distributions are inherently diffused across open vocabularies, synonymous expressions, and complex contextual formatting constraints. Consequently, token-level log probabilities from MLLMs cannot robustly represent absolute visual confidence. To ensure statistical reliability, the reported ACC represents the average result across multiple independent runs.

\begin{table*}[t]
\centering
\caption{Comparison of classification accuracy on the 12 MedMNIST 2D datasets. The best result in each column is highlighted in bold, and the second-best result is underlined.}

\fontsize{8}{9}\selectfont

\begin{tabular*}{\textwidth}{@{\extracolsep{\fill}}llcccccc@{}}
\toprule
\textbf{Category} & \textbf{Method} 
 & \shortstack{Path\\MNIST} 
 & \shortstack{Chest\\MNIST} 
 & \shortstack{Derma\\MNIST} 
 & \shortstack{OCT\\MNIST} 
 & \shortstack{Pneumonia\\MNIST} 
 & \shortstack{Retina\\MNIST} \\ 
\midrule

\multirow{5}{*}{\shortstack[l]{Supervised\\(Full-Tuning)}} 
 & ResNet50\cite{ResNet} & 0.896 & \underline{0.947} & 0.730 & 0.830 & 0.885 & 0.508 \\
 & ViT-B\cite{ViT} & 0.909 & \textbf{0.948} & 0.766 & 0.795 & 0.872 & 0.515 \\
 & Swin-B\cite{Swin} & 0.920 & \textbf{0.948} & 0.795 & 0.837 & 0.875 & 0.525 \\
 & ConvNeXt-B\cite{ConvNeXt} & 0.911 & \textbf{0.948} & 0.762 & 0.836 & 0.862 & 0.533 \\ 
 & SigLIP2-B\cite{SigLIP} & 0.941 & \textbf{0.948} & 0.818 & \underline{0.883} & 0.901 & 0.556 \\ \midrule

\multirow{2}{*}{\shortstack[l]{General\\MLLMs}} 
 & GPT-5-mini & 0.393 & 0.914 & 0.374 & 0.440 & 0.535 & 0.350 \\
 & Gemini-3-flash & 0.649 & 0.916 & 0.621 & 0.828 & 0.846 & 0.438 \\ \midrule

\multirow{2}{*}{\shortstack[l]{Medical\\MLLMs}} 
 & LLaVA-Med\cite{15} & 0.164 & 0.728 & 0.129 & 0.214 & 0.405 & 0.118 \\
 & MedGemma\cite{16} & 0.432 & 0.609 & 0.328 & 0.255 & 0.377 & 0.148 \\ \midrule

\multirow{2}{*}{\shortstack[l]{Finetuned\\Medical MLLMs}} 
 & BiomedGPT\cite{17} & \textbf{0.956} & - & \textbf{0.862} & \textbf{0.922} & \textbf{0.936} & \textbf{0.893} \\
 & Hulu-Med\cite{18} & 0.926 & - & 0.777 & 0.860 & 0.833 & - \\ \midrule

\textbf{Proposed} & \textbf{Ours} & \underline{0.954} & 0.909 & \underline{0.836} & 0.848 & \underline{0.927} & \underline{0.575} \\ 
\bottomrule
\end{tabular*}

\begin{tabular*}{\textwidth}{@{\extracolsep{\fill}}llcccccc@{}}
\toprule
\textbf{Category} & \textbf{Method} 
 & \shortstack{Breast\\MNIST} 
 & \shortstack{Blood\\MNIST} 
 & \shortstack{Tissue\\MNIST} 
 & \shortstack{OrganA\\MNIST} 
 & \shortstack{OrganC\\MNIST} 
 & \shortstack{OrganS\\MNIST} \\ 
\midrule

\multirow{5}{*}{\shortstack[l]{Supervised\\(Full-Tuning)}} 
 & ResNet50 & 0.853 & 0.965 & 0.682 & 0.946 & 0.910 & 0.789 \\
 & ViT-B & 0.814 & 0.970 & 0.665 & 0.900 & 0.895 & 0.754 \\
 & Swin-B & 0.808 & 0.977 & \underline{0.706} & 0.941 & 0.918 & 0.790 \\
 & ConvNeXt-B & 0.782 & 0.967 & 0.693 & 0.925 & 0.906 & 0.764 \\ 
 & SigLIP2-B & 0.891 & \underline{0.984} & 0.702 & \underline{0.956} & \underline{0.943} & \textbf{0.802} \\ \midrule

\multirow{2}{*}{\shortstack[l]{General\\MLLMs}} 
 & GPT-5-mini & 0.718 & 0.306 & 0.136 & 0.207 & 0.185 & 0.149 \\
 & Gemini-3-flash & 0.705 & 0.641 & 0.182 & 0.372 & 0.349 & 0.299 \\ \midrule

\multirow{2}{*}{\shortstack[l]{Medical\\MLLMs}} 
 & LLaVA-Med & 0.417 & 0.165 & 0.197 & 0.106 & 0.117 & 0.110 \\
 & MedGemma & 0.712 & 0.204 & 0.208 & 0.212 & 0.197 & 0.162 \\ \midrule

\multirow{2}{*}{\shortstack[l]{Finetuned\\Medical MLLMs}} 
 & BiomedGPT & 0.797 & 0.983 & - & - & 0.911 & - \\
 & Hulu-Med & \textbf{0.937} & 0.966 & - & - & 0.790 & - \\ \midrule

\textbf{Proposed} & \textbf{Ours} & \underline{0.905} & \textbf{0.987} & \textbf{0.718} & \textbf{0.962} & \textbf{0.944} & \underline{0.795} \\ 
\bottomrule
\end{tabular*}
\end{table*}

\subsection{Results}
As shown in Table 1, our Clinician Mimetic Workflow achieves an average accuracy of 86.3\% across the 12 MedMNIST 2D datasets. Without any backbone parameter updates, our method outperforms zero-shot general and medical MLLMs (+29.2\% average differences compared to Gemini-3-flash), and demonstrates comparable performance to fully supervised vision baselines and domain-specific fine-tuned medical MLLMs (-3.6\% average differences compared to BiomedGPT), consistently achieving the best or second-best results across the majority of the dataset. We note a relative performance bottleneck in multi-label classification scenarios, where capturing multiple concurrent pathologies within a limited context length remains challenging. The specific algorithmic causes for this bottleneck are detailed in the subsequent ablation study.

\begin{table*}[t]
\centering
\caption{Ablation study on the 12 MedMNIST 2D datasets. Top-K employs a fine-tuned visual feature extractor, whereas DECS$^\dagger$ utilizes an unadapted one. The best result in each column is highlighted in bold.}
\label{tab:ablation_medmnist}

\fontsize{8}{9}\selectfont

\begin{tabular*}{\textwidth}{@{\extracolsep{\fill}}lccccccc@{}}
\toprule
\textbf{Exemplar} & \textbf{SRES} 
& \shortstack{\textbf{Path}\\\textbf{MNIST}}
& \shortstack{\textbf{Chest}\\\textbf{MNIST}}
& \shortstack{\textbf{Derma}\\\textbf{MNIST}}
& \shortstack{\textbf{OCT}\\\textbf{MNIST}}
& \shortstack{\textbf{Pneumonia}\\\textbf{MNIST}}
& \shortstack{\textbf{Retina}\\\textbf{MNIST}} \\
\midrule
None &  & 0.347 & 0.898 & 0.426 & 0.422 & 0.766 & 0.426 \\
None & \checkmark & 0.703 & 0.906 & 0.627 & 0.571 & 0.821 & 0.463 \\
Random &  & 0.316 & 0.895 & 0.487 & 0.389 & 0.702 & 0.378 \\
Top-K &  & 0.836 & 0.901 & 0.738 & 0.706 & 0.824 & 0.499 \\
DECS &  & 0.899 & 0.901 & 0.773 & 0.723 & 0.847 & 0.519 \\
DECS$^\dagger$ & \checkmark & 0.730 & 0.903 & 0.643 & 0.576 & 0.838 & 0.496 \\
DECS & \checkmark & \textbf{0.954} & \textbf{0.909} & \textbf{0.836} & \textbf{0.848} & \textbf{0.927} & \textbf{0.575} \\
\bottomrule
\end{tabular*}

\begin{tabular*}{\textwidth}{@{\extracolsep{\fill}}lccccccc@{}}
\toprule
\textbf{Exemplar} & \textbf{SRES} 
& \shortstack{\textbf{Breast}\\\textbf{MNIST}}
& \shortstack{\textbf{Blood}\\\textbf{MNIST}}
& \shortstack{\textbf{Tissue}\\\textbf{MNIST}}
& \shortstack{\textbf{OrganA}\\\textbf{MNIST}}
& \shortstack{\textbf{OrganC}\\\textbf{MNIST}}
& \shortstack{\textbf{OrganS}\\\textbf{MNIST}} \\
\midrule
None &  & 0.500 & 0.267 & 0.306 & 0.206 & 0.221 & 0.203 \\
None & \checkmark & 0.687 & 0.739 & 0.566 & 0.740 & 0.722 & 0.601 \\
Random &  & 0.607 & 0.221 & 0.286 & 0.213 & 0.219 & 0.196 \\
Top-K &  & 0.794 & 0.856 & 0.604 & 0.806 & 0.797 & 0.684 \\
DECS &  & 0.813 & 0.951 & 0.671 & 0.882 & 0.865 & 0.711 \\
DECS$^\dagger$ & \checkmark & 0.750 & 0.818 & 0.535 & 0.754 & 0.721 & 0.626 \\
DECS & \checkmark & \textbf{0.905} & \textbf{0.987} & \textbf{0.718} & \textbf{0.962} & \textbf{0.944} & \textbf{0.795} \\
\bottomrule
\end{tabular*}

\end{table*}

\subsection{Ablation Study}
We evaluate the individual contributions of each module in Table 2. The baseline frozen backbone MLLM achieves an average accuracy of only 41.6\%. Independently integrating DECS or SRES improves the average accuracy to 79.6\% and 67.8\% respectively. Synergizing both modules yields the highest overall performance of 86.3\%. We further investigate the impact of visual referencing strategies. Top-K ICL and the unadapted DECS$^\dagger$ yield average accuracies of 75.4\% and 69.6\%, respectively. This highlights that the framework's robust performance is intrinsically driven by the synergy between DECS and SRES, rather than visual feature fine-tuning.

However, we observe limited performance gains on the multi-label ChestMNIST dataset. Utilizing DECS alone brings only a marginal improvement of 0.3\%, as a limited number of exemplars cannot comprehensively cover the combinatorial evidence space of 14 pathologies. Concurrently, utilizing SRES alone only yields a 0.8\% improvement. Empirical log analysis reveals that the SRES evaluator exhibits reward bias towards high-frequency labels, suppressing the extraction of Diagnostic Heuristics for severely class-imbalanced conditions.

\section{Conclusion}
In this work, we propose a unified Clinician Mimetic Workflow integrating DECS and SRES to adapt frozen general-purpose MLLMs for specialized medical diagnosis. By emulating human clinical reasoning, our method outperforms zero-shot general and medical MLLMs on most MedMNIST 2D datasets, achieving competitive accuracy against fully supervised baselines and fine-tuned domain models. Future work will explore multimodal experience banks, 3D datasets, and automated clinical report generation to enhance real-world clinical utility.

%
%
%
\bibliographystyle{splncs04}
\bibliography{mybibliography}

\end{document}